# Causes in neuron diagrams, and testing causal reasoning in Large Language Models. A glimpse of the future of philosophy?

Louis Vervoort[1,*], Vitaly Nikolaev[2]

*[1] Higher School of Economics, School of Philosophy and Cultural Studies, Moscow, Russian Federation*
*[2] Russian Academy of Science, Computer Science and Control, Kaliningrad Branch, Kaliningrad, Russian Federation*

*[*]Corresponding author, email for correspondence: lvervoort@hse.ru*

30.04.2025

**Abstract**. We propose a test for abstract causal reasoning in AI, based on scholarship in the philosophy of causation, in particular on the neuron diagrams popularized by D. Lewis. We illustrate the test on advanced Large Language Models (ChatGPT, DeepSeek and Gemini). Remarkably, these chatbots are already capable of correctly identifying causes in cases that are hotly debated in the literature. In order to assess the results of these LLMs and future dedicated AI, we propose a definition of cause in neuron diagrams with a wider validity than published hitherto, which challenges the widespread view that such a definition is elusive. We submit that these results are an illustration of how future philosophical research might evolve: as an interplay between human and artificial expertise.

## 1. Introduction.

Recently, computer scientists, cognitive scientists and others have systematically tested Large Language Models (LLMs) as ChatGPT[1] on a wide series of different reasoning skills, sometimes with truly impressive results. A partial list of examples includes logical reasoning (e.g. Liu et al. 2023, Bang et al. 2023), mathematical reasoning (e.g. Frieder et al. 2023, Wardat et al. 2023), physical reasoning (e.g. Lehnert 2023, West 2023, Zhang et al. 2025), psychological reasoning (e.g. Hagendorff 2023, Holterman and van Deemter 2023), medical reasoning (e.g. Bhayana et al. 2023) and several other types; this list may well grow exponentially fast in the near future (see also e.g. Bang et al. 2023, Bubeck et al. 2023, Huang and Chang 2022, Mahowald et al. 2023). From the currently best-performing LLMs, GPT-4 has been tested most extensively; two other popular LLMs that performed best on deep-reasoning tasks in physics (Zhang et al. 2025) are the open-source DeepSeek-R1 (from the company DeepSeek, cf. Liu et al. 2024a, Guo et al. 2025) and Gemini 2.0 Flash Thinking[2] (from Google DeepMind, cf. Anil et al. 2023).

---

[1] From the company OpenAI. ChatGPT, based on the Generative Pre-trained Transformer (GPT) language model, is available in different versions, such as GPT-3.5, GPT-4 (cf. OpenAI 2023). OpenAI's 'o3-mini' is a more recent model (cf. https://openai.com/index/openai-o3-mini/).

Fueled notably by the recent conviction that 'causal AI' might be a promising paradigm for next-generation systems (Peters et al. 2017, Pearl 2019), the computer science community has also started testing the causal reasoning capacities of LLMs (e.g. Gao et al. 2023, Jin et al. 2023, Kiciman et al. 2023, Tu et al. 2023, Zečević et al. 2023, Liu et al. 2024, Wang 2024). Causal reasoning is often considered a necessary ingredient of artificial general intelligence (AGI). As computer scientist and philosopher Judea Pearl puts it: "Machines' lack of understanding of causal relations is perhaps the biggest roadblock to giving them human-level intelligence" (Pearl 2019); a verdict one also finds in a philosophical analysis of deep learning (Buckner 2024, p 74). Now, causal reasoning presupposes an understanding of what causes (and effects) are; and causation is surely among philosophy's all-time favorite topics. Yet, the precise definition of cause is highly debated in philosophy. Many, and perhaps most, believe that an overarching definition of cause is illusory. L. A. Paul and Ned Hall phrase it this way, at the end of their reference work on causation (2013, p. 249): "After surveying the literature in some depth, we conclude that, as yet, there is no reasonably successful reduction of the causal relation. And correspondingly, there is no reasonably successful conceptual analysis of a philosophical causal concept. […] Barring a fundamental change in approach, the prospects of a relatively simple, elegant and intuitively attractive, unified theory of causation, whether ontological reduction or conceptual analysis, are dim".

In this context, one may wonder whether AI research and philosophical expertise on causation might fruitfully interact. Two relevant questions, the first ambitious and yet speculative, the second more topical, are: Q1) Could present or future AI help in constructing an overarching concept of cause?; and Q2) What is the status of the causal reasoning skills of a given AI, say DeepSeek or ChatGPT, notably in subtle cases – considered the preserve of philosophy? A more pragmatic variant of Q2) is Q3): How to develop a test for causal reasoning in AI based on philosophical expertise? Here we will focus on Q3); but we hope to show that advanced LLMs have made Q1) a much more pressing question than often assumed. Thus, our first main objective is to propose a testing method that is based on scholarship in neuron diagrams, widely used as heuristic tools in causation research. An ancillary objective is to illustrate the method on a few chatbots, namely ChatGPT, Gemini 2.0 Flash, and DeepSeek-R1 (which do best in certain advanced tests, cf. Zhang et al. 2025). An existing test for *abstract* causal reasoning developed by computer scientists assesses whether an LLM can infer causation from a list of correlations (Jin et al. 2023). Here we propose a test that is simpler in application and that directly probes whether the LLM can identify causes in various abstract scenarios, notably in subtle cases involving redundant causation, causation by omission, violation of transitivity, etc. In view of the likely rise of ever more powerful AIs, in particular 'causal AI' (Pearl 2019, see also Hartnett 2018, Zečević et al. 2023), it seems important to have a variety of tests addressing various causal reasoning skills. In this context, philosophers have recently proposed a roadmap for developing tests for scientific understanding in LLMs (Barman et al. 2024). For a critical philosophical assessment of the capacities of GPT-4 in general reasoning, see (Arkoudas 2023, 2023a, Floridi 2023).

---

[2] Cf. https://ai.google.dev/gemini-api/docs/thinking

As said, even if we focus on Q3), Q1) came in through the back door. The surprising causal reasoning skills of certain LLMs forced us to propose a definition of cause, one that is more encompassing than the definitions proposed in the literature. To construct a more general definition applicable to neuron diagrams, then, is the second objective of this article; it led, *in fine*, to our main theoretical result. Such a definition was necessary to verify the correctness of the LLM's answers to questions of the type: "What is the cause of event E in neuron diagram D?" Here we propose a definition of cause that is in agreement with intuition (and with the verdicts that can be found in the literature) for all the neuron diagrams we studied, which is a sizeable part of the diagrams discussed in the reference (Paul and Hall 2013). To the best of our knowledge, a definition with such a broad domain of validity has not yet been constructed. Hence, the interaction with advanced AI was at the origin of a new philosophical result, in an area that draws vast theoretical and practical attention. Since it is reasonable to conjecture that the capacities of AIs will increase in the near future, we submit that a constructive interaction between human and artificial expertise will soon become a reality also in theoretical philosophical research, as already hinted to by the present work.

A word about our background assumptions. We will not delve here into the question whether LLMs can 'really' reason causally. Our research resonates best with a pragmatic conception of intelligence, centered on problem solving, which has become popular in AI since Turing (cf. Norvig and Russell 2015, Bringsjord and Govindarajulu 2018). In this perspective, the distinction between 'mastering a type of reasoning' and 'emulating/simulating' this type of reasoning is of secondary importance: what counts is the capacity to solve problems and answer questions – for instance of the type 'what causes X?' In sum, we remain here largely agnostic of whether the skill to systematically answer questions and solve problems in a certain cognitive field (at a defined level of proficiency) is indicative of 'really' mastering the corresponding skill (at that level of proficiency).

While our testing method could be used to test any LLM (and other AIs), we will illustrate the proposed method on advanced versions of ChatGPT, DeepSeek and Gemini. These experiments mainly aim at illustrating the principle and feasibility of the testing method, and at suggesting lines of further research. While our preliminary test results already show instances of impressive causal reasoning skills of these LLMs, they were not obtained by large-scale statistical experiments and quantitative methods as they are deployed in computer science, cognitive science, psychology etc. to come to, ideally, objective conclusions (for an example of a comprehensive benchmark test, BIG-bench, see Srivastava et al. 2022). Yet, our test could be further developed for such a systematic inquiry[3], and possibly for assessing and comparing the depth-of-reasoning of various AIs, as we suggest in Section 4. For our own proof-of-concept tests, we first used ChatGPT based on GPT-3.5 (in July 2023) and GPT-4 (in July 2023 and March 2024), and finally DeepSeek-R1 (DeepThink), Gemini 2.0 Flash (Thinking Experimental), and ChatGPT o3-mini (in February 2025). In the following we refer to the first two versions of ChatGPT as ChatGPT(3) and ChatGPT(4), respectively.

The article is organised as follows. In Section 2 we first give a succinct overview of testing methods and results obtained in the computer science community. Then we describe our test, after an explanation of how neuron diagrams work. We show the results of a small-scale experiment with ChatGPT(4) (and more results in Appendix 1); comparison with the other chatbots is given in Appendix 2. In Section 3 we propose a definition of cause that allows one to derive the intuitive causes for the 'classic' neuron diagrams used in our test, and that can therefore serve as the 'gold standard' (as it is called in computer science) to assess the answers by LLMs. Such a definition does not apply to literally *all* causal scenarios and corresponding diagrams studied in the literature, but is more encompassing than other proposed definitions and should therefore offer a solid basis for further synthesis. Section 4 is devoted to a discussion of the test results and to lines of further (interdisciplinary) research. Section 5 concludes.

## 2. Test for abstract causal reasoning, and results obtained by ChatGPT and other LLMs

Before turning to our test, let us briefly comment on related work by computer scientists, who have performed several tests on causal reasoning in LLMs and ChatGPT (Liu et al. 2024, Gao et al. 2023, Jin et al. 2023, Kiciman et al. 2023, Tu et al. 2023, Zečević et al. 2023, Wang 2024). Computer scientists' assessments of ChatGPT's capabilities as 'causal reasoner' vary greatly (even assuming that the test results have statistical significance): from enthusiastic (Kiciman et al. 2023) to much more pessimistic (Jin et al. 2023, Zečević et al. 2023). An obvious reason for this discrepancy is that many different cognitive skills have been classified as forms of causal reasoning; and different types of tests measure different skills[4]. Among the enthusiasts, Kiciman et al. (2023) conclude: "We envision LLMs to be used alongside existing causal methods, as a proxy for human domain knowledge and to reduce human effort in setting up a causal analysis, one of the biggest impediments to the widespread adoption of causal methods." Gao et al. (2023), who claim to have conducted the first comprehensive evaluation of ChatGPT's causal reasoning capabilities, conclude that ChatGPT is not a good 'causal reasoner' (cannot reliably identify causes in concrete situations), but a good 'causal explainer' (can reliably come up with plausible explanations of why a causal relation exists in a concrete situation), while having a serious hallucination problem with causal reasoning[5]. Zečević et al. (2023) come to the conclusion that LLMs are only 'causal parrots' that cannot reason causally but only textually reproduce causal links they have learned during training.

---

[3] Whether this is useful could be explored by an interdisciplinary research effort with computer scientists.

A key demarcation criterion in these different types of tests is whether they focus on *concrete* causal situations (described in texts), i.e. referring to particular real-world facts, events, scenarios, or on *abstract* scenarios, i.e. described in a formalised manner using variables. The majority of the existing tests is of the first type. Jin et al. (2023) have proposed the first benchmark dataset to test abstract causal inference skills of LLMs, using, notably, the work on causal discovery by Spirtes, Glymour and Scheines (2000). Their test aims at assessing whether an LLM can extract from the list of correlations that exist between variables the correct causal graph, for the time being limited to maximum six nodes (i.e., variables). Their test results for this particular causal skill are sobering, and show that LLMs, including ChatGPT(4), do not perform better than random guessing. (Even after finetuning these models fail to generalize: they fail in 'out-of-distribution' settings, i.e. in queries where variable names and textual expressions are not similar to those seen in the training set.)

---

[4] As examples, Gao et al. (2023) include, besides causal identification as we study below, discovering causal relationships and providing explanations of causal facts. Kiciman et al. (2023) mention causal inference from data (covariance analysis); causal inference by logical, e.g. counterfactual, reasoning; discovering causal graphs; inference of event versus type causality; of normal versus contrived causality; of causal direction; etc.

[5] These authors also claim that the data set of e.g. Kiciman et al. (2023) was too small, leading to an overestimation of the causal reasoning skills.

The complementary testing method we describe now is a test of abstract causal reasoning, directly inspired by research in the philosophy of causation, so focusing on scenarios with a certain degree of complexity and/or subtlety. The principle of the method is straightforward, and lends itself to systematization. The method also could, in principle, be enriched (cf. Section 4) and used to develop a large set of testing scenarios (in a benchmark spirit), since large neuron diagrams can easily be generated by computer code (so one can easily go beyond the 6-node limit of Jin et al. 2023).

In summary, the test we propose assesses whether the AI under scrutiny is able, when presented with the textual version of a neuron diagram, to give the correct answer to questions of type Q-CAUSE: "what is the cause of event E?" In our tests we added one more question, namely "does event E occur?", where E typically is the firing of the last neuron in a diagram (cf. examples below). Clearly, being able to answer Q-CAUSE with sufficient proficiency (in a statistically meaningful ensemble of causal situations) can be considered a measure of (a certain type of) abstract causal reasoning.

In order to answer questions as Q-CAUSE in subtle cases, when 'automatic' or 'implicit' intuition is uncertain, one would like to rely on an explicit definition of cause. One prominent model is the counterfactual interpretation of cause, which is based on following sufficient condition (Lewis 1973): (Actual) event C is a cause of (actual) event E if the following holds: if C were not occur, E would not occur. We will call this in the following the 'simple counterfactual condition/rule'. Starting from this condition a great number of counterfactual definitions have been proposed, many of them discussed at length in (Paul and Hall 2013). The limits of such definitions or models have most efficiently been studied by so-called neuron diagrams, popularised by David Lewis; presumably the largest collection of those can be found in (Paul and Hall 2013), which we use here to extract our sample basis. Neuron diagrams allow one to represent a wide variety of causal situations and problems discussed in the literature, including causal redundancy ('early/late preemption', overdetermination), causal omissions, transitivity, etc. Let us briefly explain how they work.

A representative diagram is given in Fig. 1, reproduced from (Paul and Hall 2013).

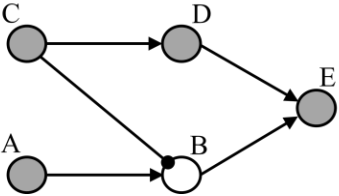

Fig. 1. Typical neuron diagram (reproduced from Fig. 1, Paul and Hall 2013, p. ix).
'Firing' or 'on' neurons are shaded.

In the diagram of Fig. 1 five subsystems, represented by five neurons, are in causal interaction in the following way. Neuron C emits a stimulating signal (represented by an arrow) towards neuron D, which subsequently fires; D likewise sends a stimulating signal to E (all firing/'on' neurons are shaded). A stimulates B but B is inhibited to fire due to the inhibitory signal from C, represented by a line with a black dot[6]. The temporal order goes from left to right: C and A fire at t1, B and D react at t2, E reacts at t3, where t1 < t2 < t3. Note that the (causal) connections, so the functioning or meaning of the diagram, can be described without using the word 'cause'. (When no confusion can arise, we will also indicate the *firing* of neuron X with the same symbol X, which therefore can stand for the neuron or the event. If needed for clarity we will use 'X-neuron'.) Importantly, it is on 'classic' diagrams defined by the rules just given that we will base our test[7]. In Paul and Hall 2013 other, sometimes more complex, types of diagrams are discussed, which need additional or different rules of functioning.

Actually, Fig. 1 represents one of the most discussed cases in the recent analytical literature, called 'early preemption'. Looking at the diagram, the intuitive verdict is that the cause(s) of E's firing are the firing of C and/or D. Or if one wishes to be more precise, one could state that C is the 'root' cause at t1 and D the 'proximate' cause at t2. Intuitively the firing of A would *not* be identified as a cause, because its action is blocked by C. But this intuitive identification of C as the cause of E conflicts with the simple counterfactual condition just given: it seems clear that if C would not happen, E would still happen, due to the back-up neuron A.

Conflicts with intuition as the above have been detected for a wide variety of diagrams, which are hotly debated. No encompassing definition of cause has been identified in the literature, one that works for all (classic or other) neuron diagrams, and ideally for all causal situations; there is a widespread consensus that such a universal definition cannot be constructed. At the same time, human intuition converges for very many causal situations – indeed, historically these converging intuitions have been the standard by which the neuron diagrams have been assessed, and candidates for an encompassing definition disqualified.

For our purposes we will not go in any detail in the vast literature devoted to defining cause in neuron diagrams. What matters to us is that the expertise gathered on causation can be put to use in a pragmatic way, namely by using the diagrams to test the causal reasoning capacity (or its simulation/emulation) of a given AI. In the case of LLMs as ChatGPT this is straightforward, since the diagrams can be textually transcribed. The full series of 25 diagrams with which we tested ChatGPT is given in Table 1 in Appendix 1. These include a part of the cases discussed in (Paul and Hall 2013), as well as several variations of these diagrams; in (Paul and Hall 2013) about 50 diagrams are discussed. We selected the diagrams of Table 1 essentially for following reasons: 1) they are classic in the sense we defined, i.e. based on the same rules of functioning as those of Fig. 1, and can (therefore) easily be transcribed; 2) the ensemble is, roughly, representative of the degree of complexity of the collection in (Paul and Hall 2013); 3) they allowed for constructing variations that were helpful for identifying and verifying our new definition DEF-1 (cf. Section 3). Again, we have no pretence at completeness[8].

---

[6] In the words of Paul and Hall: "Arrows represent token-level stimulatory connections between neurons; lines ending with black dots represent token-level inhibitory connections" (2013, p. 10).

[7] For a few diagrams of Tables 1 and 2, such as diagram 10, one more rule of functioning is needed: if a neuron has a double border, it only fires upon reception of at least two stimulating signals.

A sub-ensemble of detailed results obtained on ChatGPT(4) is given in Table 2 below, which also illustrates how we did the transcription (column 2). As an example, the transcription of diagram 1 in Tables 1-4 and Fig. 1, and the questions we ask, are as follows: "Suppose time t1 is earlier than time t2, which is earlier than time t3. If C would occur at t1, D would occur at t2. If D would occur at t2, E would occur at t3. If A would occur at t1, B would occur at t2, unless C would occur at t1. If B would occur at t2, E would occur at t3. Suppose C and A occur at t1. Does E occur at t3? What is/are the cause(s) of E's occurring or not occurring?"

So we always ask essentially the same question, namely whether the last neuron in the diagram fires or not, and what the causes of the firing or not firing are. (In order to guarantee a certain stability and representativeness, the experiments ran over a period of several months in 2023 and again at the beginning of 2024, first with ChatGPT(3) then with ChatGPT(4), on which we focus here. The answers shown in the tables were obtained in one single run (lasting less than an hour, and without feedback to ChatGPT), and are relatively stable over time: the replies and therefore the small-scale statistics of correct/incorrect answers did not change much. In sum, the results shown here should be representative enough for our philosophical analysis.)

Table 2 (and subsequent tables) also shows the 'correct answers' in column 3 and the answers given by ChatGPT(4) in column 4. Let us be clear about how we define 'correct answers' in the tables. Correct answers are (i) whenever the diagram is discussed in (Paul and Hall 2013), those that are considered the intuitive answers by these authors; (ii) when the diagram is not discussed there, the intuitive answers according to our personal judgement; (iii) *in all cases*, the causes identified by DEF-1 (cf. next section). In sum, we believe that for all, or almost all[9], answers intuitions can converge, at least with a minimal training. Most importantly, the causes indicated in the tables can be derived with one definition, DEF-1 below, to which we turn in the next section.

---

[8] Several 'non-classic' diagrams from (Paul and Hall 2013) can also be transcribed in text; we used a few for testing ChatGPT, with results to be published elsewhere. The ratio correct/incorrect answers was quite comparable to the one found in the present study.

Note that in unambiguous cases we do not always indicate *all* causes in the tables. For instance, for diagram 5 we only indicate as causes of E's occurring: A+(t1); D+(t2); F–(t3). Here A+(t1) stands for 'the firing/occurrence of A at time t1'; F–(t3) for the 'the non-firing/non-occurrence of F at t3'; etc. The intermediate causes between A+(t1) and E+(t4), namely A1+(t2) and A2+(t3), are obvious once A+(t1) is identified as a root or initial cause (we justify this point in the next section). The corresponding intermediate neurons A1 and A2 are not labelled in the diagram (but of course mentioned in the transcription); a similar simplification is made in other diagrams. Thus, we only indicate the most relevant/'difficult' causes, indeed those that are discussed in the literature, typically those occurring at t1. In diagram 5 for instance, representing a case of double prevention, the most relevant feature is that usually only A+(t1) is counted as a root cause, not C+(t1), which seems to violate transitivity (Paul and Hall 2023, p. 224).

Indeed, as we will corroborate in the next section, the initial causes of the last event in the diagrams are often least easy to identify. As follows from intuition *and* DEF-1 below, proximate causes of a given effect, i.e. causes that are only one step earlier in time, are always (comparatively easily) identified by the simple counterfactual rule mentioned above. For instance, in Fig. 1, this rule immediately allows one to find that D+ (the firing of neuron D) is a cause at t2 of E+ at t3, and that B- (the non-firing of B) is *not* a cause of E+. Using the same counterfactual rule one identifies in diagram 5 (Table 2) A2+ and F- as proximate causes of E+, D+ of F-, and C+ of D+ and B+. So proximate causes are easily identified by the simple counterfactual rule.

| Diagram | Transcription + Question | Correct answer | ChatGPT(4)'s answer |
|---|---|---|---|
| **1.** (From PH, Fig. 1, p. ix) | Suppose time t1 is earlier than time t2, which is earlier than time t3. If C would occur at t1, D would occur at t2. If D would occur at t2, E would occur at t3. If A would occur at t1, B would occur at t2, unless C would occur at t1. If B would | Yes. **C+(t1); D+(t2)** | If both A and C occur at t1, then according to your stipulations, at t2, D would occur (caused by C) and B would not occur (because A at t1 prevents B from occurring at t2). Since B does not occur at t2, the condition for E to occur at t3 based on B is not met. However, D does occur at t2, which, according to |
| 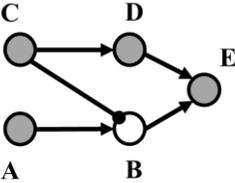 | occur at t2, E would occur at t3. Suppose C and A occur at t1. Does E occur at t3? What is/are the cause(s) of E's occurring or not occurring? | | your conditions, does cause E to occur at t3. So, E does occur at t3, and the cause is D occurring at t2, which in turn is caused by C occurring at t1. *<CORRECT>* |
| **2.**<br>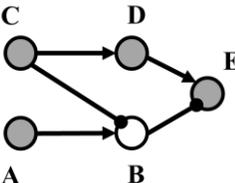 | Suppose time t1 is earlier than time t2, which is earlier than time t3. If C would occur at t1, D would occur at t2. If D would occur at t2, E would occur at t3 unless B would occur at t2. If A would occur at t1, B would occur at t2, unless C would occur at t1. Suppose C and A occur at t1. Does E occur at t3? What is/are the cause(s) of E's occurring or not occurring? | Yes.<br>**C+(t1); D+(t2), B-(t2)** | Given the conditions and that C and A occur at t1, we can deduce the following:<br>1. Since C occurs at t1, D will occur at t2.<br>2. The occurrence of C at t1 means that even though A also occurs at t1, B will not occur at t2 (as per the rule "If A would occur at t1, B would occur at t2, unless C would occur at t1").<br>3. Since D occurs at t2 and B does not, there is nothing to prevent E from occurring at t3. Therefore, E does occur at t3.<br>The cause(s) of E's occurring is the presence of C and A at t1 which leads to the presence of D at t2, and the non-occurrence of B at t2 (due to the presence of C at t1), allowing E to occur at t3.<br>*<PARTLY CORRECT>* |
| **5.** (From PH, Fig. 10, p. xi)<br>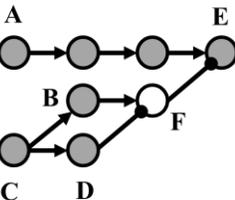 | Suppose time t1 is earlier than time t2, which is earlier than time t3, which is earlier than time t4. If A would occur at t1, A1 would occur at t2. If A1 would occur at t2, A2 would occur at t3. If A2 would occur at t3, E would occur at t4, unless F occurs at t3. If C would occur at t1, B and D would occur at t2. If B would occur at t2, F would occur at t3, unless D would occur at t2. Suppose both C and A occur at t1. Does E occur at t4? What is/are the cause(s) of E's occurring or not occurring? | Yes.<br>**A+(t1); D+(t2); F-(t3)** | Given the conditions, E does not occur at t4.<br>Here's how the events unfold:<br>1. At t1, both A and C occur.<br>2. At t2, as a result of A's occurrence at t1, A1 occurs. As a result of C's occurrence at t1, both B and D occur.<br>3. At t3, A2 occurs because A1 occurred at t2. However, F does not occur because although B occurred at t2, D also occurred at t2, which prevents F from occurring.<br>4. At t4, E does not occur because even though A2 occurred at t3, F did not occur at t3.<br>So, the primary reason for E not occurring at t4 is the non-occurrence of F at t3, which is due to the occurrence of both B and D at t2, which in turn was caused by C's occurrence at t1.<br>*<INCORRECT>* |
| **8.** (From PH, Fig. 12, p. xii)<br>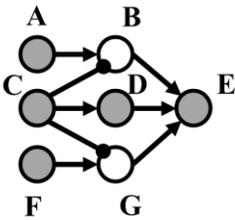 | Suppose time t1 is earlier than time t2, which is earlier than time t3. If A would occur at t1, B would occur at t2, unless C would occur at t1. If B would occur at t2, E would occur at t3. If C would occur at t1, D would occur at t2. If D would occur at t2, E would occur at t3. If F would occur at t1, G would occur at t2, unless C would occur at t1. If G would occur at t2, E would occur at t3. Suppose A, C, and F all occur at t1. Does E occur at t3? What is/are the cause(s) of E's occurring or not occurring? | Yes.<br>**C+(t1); D+(t2)** | Given the conditions, E does occur at t3.<br>Here's how the events unfold:<br>1. At t1, A, C, and F all occur.<br>2. At t2, as a result of C's occurrence at t1, D occurs. However, B does not occur because A's effect was negated by the presence of C, and G does not occur because F's effect was also negated by the presence of C.<br>3. At t3, E does occur, as a result of D's occurrence at t2.<br>So, the primary reason for E's occurrence at t3 is the occurrence of D at t2, which was caused by C's occurrence at t1.<br>*<CORRECT>* |
| **10.** (From PH, Fig. 14, p. xiii) | Suppose time t1 is earlier than time t2, which is earlier than time t3. If A would occur at t1, B would occur at t2. If A would occur at t1, D would occur at t2, unless C would occur at t1. If B would occur at t2, E would occur at t3, if also either D at t2 or F at t2 (or both) would occur. If C would occur at t1, F would occur at t2, and D would not occur at t2. If F would occur at t2, E would occur at t3 if also either D or B (or both) would | Yes.<br>**A+(t1), C+(t1); B+(t2), F+(t2)** | If both A and C occur at t1, then at t2, according to your stipulations:<br>1. B would occur because A has occurred at t1.<br>2. D would not occur because C occurring at t1 prevents D from occurring at t2.<br>3. F would occur because C has occurred at t1.<br>Given these outcomes, we then look at the conditions for E occurring at t3. You stated that E will occur at t3 if B, D, or F (or any combination of them) occur at t2. |

| | | | | |
|---|---|---|---|---|
|  | occur at t2. If D would occur at t2, E would occur at t3 if also either B or F (or both) would occur at t2. Suppose both A and C occur at t1. Does E occur at t3? What is/are the cause(s) of E's occurring or not occurring? | | In this situation, both B and F occur at t2. Hence, even without D occurring at t2, the condition for E to occur at t3 is satisfied. Therefore, E does occur at t3. The causes of E's occurrence are B and F occurring at t2, which were respectively caused by A's occurrence at t1 and C's occurrence at t1. <br><br>***<CORRECT>*** |
| **12. (From PH, Fig. 29, p. xvii)** <br>  | Suppose time t1 is earlier than time t2, which is earlier than time t3. If A would occur at t1, A1 would occur at t2. If A1 would occur at t2, E would occur at t3, unless D would occur at t2. If B would occur at t1, D would occur at t2, unless C would occur at t2. Suppose A, B and C all occur at t1. Does E occur at t3? What is/are the cause(s) of E's occurring or not occurring? | Yes. <br> **A+(t1); C+(t1)** | Given the conditions and that A, B, and C all occur at t1, we can deduce the following: <br> 1. Since A occurs at t1, A1 will occur at t2. <br> 2. As B occurs at t1 and C also occurs at t1, according to the conditions, D will not occur at t2 (since the occurrence of C at t1 prevents D from occurring at t2 even though B occurs at t1). <br> 3. Since A1 occurs at t2 and D does not occur at t2, E will occur at t3 (as the condition preventing E's occurrence at t3 - D at t2 - is not met). <br> So, E does occur at t3. The cause of E's occurring is the occurrence of A at t1 (which leads to A1 at t2 and thus to E at t3) and the occurrence of C at t1 (which prevents D from occurring at t2 and therefore removes the barrier to E occurring at t3). <br><br>***<CORRECT>*** |
| **15. (From PH, Fig. 42, p. xx)** <br>  | Suppose time t1 is earlier than time t2, which is earlier than time t3, which is earlier than time t4. If A would occur at t1, A1 would occur at t2. If A1 would occur at t2, A2 would occur at t3. If A2 would occur at t3, E would occur at t4, if also F would occur at an earlier time. If G would occur at t1, B would occur at t2, unless C would occur at t2. If B would occur at t2, F would occur at t3. If C would occur at t1, D would occur at t2. If D would occur at t2, F would occur at t3. Suppose A, G and C occur at t1. Does E occur at t4? What is/are the cause(s) of E's occurring or not occurring? | Yes. <br> **A+(t1), C+(t1); etc.** | Given the conditions and that A, G, and C occur at t1, we can deduce the following: <br> 1. Since A occurs at t1, A1 will occur at t2. Following this chain, A2 will occur at t3. <br> 2. As C occurs at t1, D will occur at t2. Following this chain, F will occur at t3. <br> 3. The occurrence of G at t1 would usually lead to B at t2, but since C also occurs at t1, the occurrence of B at t2 is prevented. <br> 4. The occurrence of F at t3 fulfills the condition for E to occur at t4 (given that A2 also occurs at t3). <br> Therefore, E does occur at t4. The cause of E's occurrence is the sequence A -> A1 -> A2 and F occurring at an earlier time, triggered by the sequence C -> D -> F. <br><br>***<CORRECT>*** |
| **17. (From PH, Fig. 43, p. xxi)** <br>  | Suppose time t1 is earlier than time t2, which is earlier than time t3, which is earlier than time t4. Only if F would occur at t1, F1 would occur at t2. Only if F1 would occur at t2, F2 would occur at t3. Only if F2 would occur at t3, G would occur at t4, unless E would occur at an earlier time. If C would occur at t1, D would occur at t2. If D would occur at t2, E would occur at t3. If A would occur at t1, B would occur at t2, unless C would occur at an earlier time. If B would occur at t2, E would occur at t3, unless C would occur at t1. Suppose F, C and A occur at t1. Does G occur at t4? What is/are the cause(s) of E's occurring or not occurring? | No. <br> **C+(t1); D+(t2); E+(t3)** | Given the conditions and that F, C, and A occur at t1, we can deduce the following: <br> 1. Since F occurs at t1, F1 will occur at t2. Following this chain, F2 will occur at t3. Theoretically, this could lead to G occurring at t4, unless E occurs at an earlier time. <br> 2. Since C occurs at t1, D will occur at t2. This chain leads to E occurring at t3. <br> 3. Since A occurs at t1 and C occurred earlier at t1, B won't occur at t2. This means there's no alternative route for E to occur at t3. <br> Therefore, given these conditions, E does occur at t3 due to the chain C -> D -> E. Because E occurs at t3, which is earlier than t4, G does not occur at t4. The cause of E's occurrence is the sequence C -> D -> E. <br><br>***<CORRECT>*** |
| **18. (From PH, Fig. 44, p. xxi)** | Suppose time t1 is earlier than time t2, which is earlier than time t3, which is earlier than time t4, which is earlier than time t5. If H would occur at t1, H1 would occur at t2. If H1 would occur at t2, H2 | Yes. <br> **H+(t1), C+(t1); H3+(t4), G-(t4); etc.** | Given the conditions and the events at t1 (H, F, C, A), let's follow the chain of events: <br> 1. H occurs at t1, so H1 will occur at t2. Then, H1 leads to H2 at t3, H2 to H3 at t4, and without interference, H3 would lead to I at t5. |
|  | would occur at t3. If H2 would occur at t3, H3 would occur at t4. If H3 would occur at t4, I would occur at t5, unless G would occur at an earlier time. If F would occur at t1, F1 would occur at t2. If F1 would occur at t2, F2 would occur at t3. If F2 would occur at t3, G would occur at t4, unless E would occur at an earlier time. If C would occur at t1, D would occur at t2. If D would occur at t2, E would occur at t3. If A would occur at t1, B would occur at t2, unless C would occur at an earlier time. If B would occur at t2, E would occur at t3. Suppose H, F, C and A occur at t1. Does I occur at t5? What is/are the cause(s) of I's occurring or not occurring? | | 2. F occurs at t1, so F1 will occur at t2. Then, F1 leads to F2 at t3, and without interference, F2 would lead to G at t4.<br>3. C occurs at t1, so D will occur at t2. Then, D leads to E at t3.<br>4. A occurs at t1. But since C also occurs at t1, B will not occur at t2 (as per the rule "If A would occur at t1, B would occur at t2, unless C would occur at t1"), so there's no additional path to E from here.<br><u>Following the sequence, the conditions indicate that G should occur at t4 (as F2 is expected at t3 and there's no E prior to t3 to prevent it). Since G happens at an earlier time than t5, this will prevent I from occurring at t5.</u><br><u>So, I does not occur at t5. The cause of I's non-occurrence is the event sequence initiated by F at t1, which leads to G occurring at t4 (before t5). Without the occurrence of F at t1, the sequence leading to G (and thus preventing I) wouldn't happen.</u><br><br>***\<INCORRECT\>*** |

**Table 2**. ChatGPT(4)'s answers (column 4) to causal questions based on neuron diagrams (see full table in Appendix 1). Column 2 gives the transcriptions and questions and column 3 the correct answers (here C+(t1) stands for 'the firing/occurrence of C at time t1'; etc.). In the last column, incorrect parts are underlined. In *\<BRACKETS\>* our assessment of the (in)correctness of the answer as a whole. All diagrams except 2 are discussed in (Paul and Hall 2013) (PH in the table). In diagram 10, the last neuron has a double border, meaning that it only fires upon reception of at least two stimulating signals. In diagram 5, the intermediate neurons A1, A2 between A and E are not labelled for simplicity; same simplification in other diagrams.

The last column of Table 2 indicates in brackets our verdict 'correct' or 'incorrect', as well as which part of ChatGPT(4)'s answer is incorrect, by underlining it. Note that in Tables 1 and 2 our evaluation <(IN)CORRECT> refers to the question as a whole, in particular the identification of causes. The diagrams selected in this table are particularly interesting, since they are all, except one, discussed in detail in (Paul and Hall 2023), as indicated in the table. The proportion of correct answers in Table 2 is quite large (6/9), larger than in the full ensemble of Table 1, where it is 13/25 (as again shown in the table).

In some detail, the results we obtained were the following. For all diagrams except 5 and 18, ChatGPT(4) correctly answered the 'yes/no'-question whether the last neuron fires or not, a much better result than obtained with ChatGPT(3), which came close to a random-guess result (50% correct if only yes/no can be answered). For the essential part of the test, focusing on finding causes, the results were less univocal. As shown in Table 1, we judged that ChatGPT(4) correctly identified the causes for 13 of the 25 submitted diagrams (for two more diagrams the answer was partly correct). ChatGPT(3) found for only one diagram (20) the correct causes. This significant increase in capability due the parametric upscaling in GPT-4 is generally seen in tests; with further upscaling of the LLM its scores might well improve.

We will comment on these results in Section 4, but since the statistics is small, we should be cautious in drawing conclusions. Let us, for now, only highlight three observations. First, the tables illustrate that Q-CAUSE works as a prompt to elicit answers and causal reasoning, or its emulation, from ChatGPT. ChatGPT(4)'s answers always seem to make perfect sense, at least on the surface. But identifying causes is, in principle, an all-or-nothing task: overlooking causes can have dramatic consequences – as any health care worker, engineer, or detective might tell. Therefore, our test is specific enough to be quantitatively graded. On the other hand, we find it remarkable that in several instances ChatGPT(4) provides answers to causal questions that are correct according to human intuition and the criterion stipulated above, even in cases where the 'correct answer', or at least its justification, is subject to debate in philosophy. This is for instance the case for the six correct answers in Table 2. Finally, we noted that for the more complex diagrams 22-25 in Table 1, introducing interrupted paths, or crossings between rows – features that are absent in the classic diagrams of (Paul and Hall 2013) –, the causal reasoning of ChatGPT(4) appeared to break down. The answers were of a saliently lower quality than for simpler diagrams. This suggests that there is a limit in complexity in the diagrams above which the causal reasoning capability of ChatGPT(4) collapses.

Qualitatively comparable results were obtained with advanced LLMs, all recently released, namely DeepSeek-R1, Gemini 2.0 Flash (Thinking Experimental) and OpenAI o3-mini, as shown in Appendix 2, Table 3. The 25 diagrams and the questions are the same as in Table 1, except that (i) we only ask to give the causes at the earliest time (t1), and (ii) we specify that the answers should be short, in order to avoid needlessly lengthy answers providing great detail on the reasoning steps (see details in Appendix 2). For this small-scale test, best-in-class was Gemini 2.0 Flash (Thinking Experimental) with 14 fully correct and 9 partially correct answers, and 2 wrong answers, followed by DeepSeek-R1 (resp. 10 / 15 / 0). ChatGPT o3-mini had only 2 fully correct and 11 partly correct answers, while 12 answers were wrong, as can be read in the Table. This last result is surprising since the older ChatGPT(4) did better. We have no compelling explanation for this phenomenon[10].

Before commenting further on these results, let us first derive a definition that works for all diagrams in the Tables (1-4). From the perspective of theoretical philosophy, this is our main result.

---

[10] Perhaps this result is due to the fact that we ask now to give short answers (we asked to not list reasoning steps) and to mention only initial causes. On the other hand, the answers of all the chatbots we studied were quite robust under variations in the precise phrasing of the prompt (cf. Section 4); but more systematic work may be needed here.

### 3. Analytic definition of cause for the neuron diagrams of Table 1.

For all the cases of Table 1 and all subsequent tables, we identified the causes X of event Y (Y is the firing or non-firing of the last neuron in the diagrams, usually labelled E) by following definition, applied to a given neuron diagram:

X is a cause of Y    *iff*    ¬X (ceteris paribus, off-path under max blocking) implies ¬Y.        (DEF-1)

This definition works for any Y-event in a diagram, not only occurring at the last neuron. It is implicitly assumed that 1) Y does not occur earlier than X – in accordance with intuition, and 2) that there is some path, i.e. a suite of contiguous stimulating or blocking connections from X to Y. In DEF-1 '¬X' can be read 'not-X' (event X not happening, corresponding to inverting 'on' and 'off' states of the X-neuron). (X and Y stand in principle for the events, firing or non-firing, but sometimes also indicate the neurons. If confusion is possible, we write e.g. 'X-neuron' or 'X-event' or 'X+' as before.)

DEF-1 needs to be made precise, as we do in the following[11]. We will show below that the synthetic phrasing of DEF-1 allows one to apply it conveniently to all special cases, once one is accustomed to the new notions ('off-path' and 'blocking') it contains. The key point is this: the interpretation of the 'off-path under max blocking' clause depends on whether X has only one forward-in-time connection, or more. In the latter case we say that X is 'bifurcating'; C in Fig. 1 is such a bifurcating neuron. In the former, much more frequent, case the clause can be neglected and one reverts to the simple counterfactual condition/rule mentioned in the previous section. Thus, it is only when bifurcating paths occur that DEF-1 needs the 'off-path under max blocking' clause.

But let us specify DEF-1 in detail; it comes with rules. To see whether X is a cause of Y, one needs to consider the factual scenario as depicted in the diagrams, and a counterfactual scenario that corresponds to the diagram in which X is replaced by ¬X and the consequences implemented in the way specified under (i) below. In this counterfactual scenario one has to evaluate whether ¬Y occurs. The counterfactual scenario is built in the following way, depending on whether the X-neuron is bifurcating or not:

(i) Both for a bifurcating or non-bifurcating X-neuron, the 'forward' consequences of the replacement X ⟶ ¬X are implemented *ceteris paribus*, so while keeping all other events, not related (i.e.

---

[11] We are aware that, a priori, DEF-1 may seem artificial or ad-hoc. Indeed, we intend to show in a forthcoming publication that a more natural and more encompassing definition than DEF-1 can be constructed. This work also allows us to precisely explain why DEF-1 is the correct definition for the 'classic' neuron diagrams we consider here.

connected) to X, fixed. Forward consequences are the 'subsequent-in-time' changes *implied by the replacement X → ¬X, according to the rules of the functioning of the diagram* (given in the description of Fig. 1). Backward consequences are neglected (one can imagine the part of the diagram occurring before X being cut off). To take the simplest example, if in the factual scenario two active neurons X and Y are only connected via a stimulating connection, ¬X will imply ¬Y in the counterfactual scenario, in which both X and Y are off. According to DEF-1, the firing of X is then a cause of the firing of Y.

(ii) If X is non-bifurcating, so if there is one path from X to Y, *or* if the X-neuron is off (whether bifurcating or not), then we do not really need[12] the 'off-path under max blocking' clause, and DEF-1 corresponds to a simple counterfactual definition (with the ceteris-paribus specification of (i)). As an example, the ceteris-paribus clause implies that in diagram 5 of Tables 1 and 2 F- is a cause of E+: F+ implies E- if one keeps fixed all F-unrelated events in the path A-E. *Most neurons are non-bifurcating, so the simple counterfactual rule is baseline for identifying causes.* Consider the case that X is off and bifurcating. Then DEF-1 just stipulates that event X (the X-neuron being off) is a cause of Y just in case ¬X implemented (in the counterfactual diagram), ceteris paribus, implies ¬Y. This case is illustrated in Fig. 2 (diagram 3 in Table 1), where two paths run from C to E.

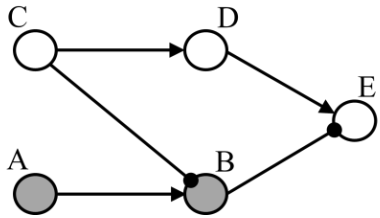

Fig. 2. The non-firing of C is a cause of the non-firing of E.

It seems intuitive that here the non-firing of C is a cause of the non-firing of E: if C would fire, E would fire – because the counterfactually firing C would block the B-neuron. This is also what DEF-1 delivers. Note that one can still consider this as a (counterfactually) 'off-path maximal blocking' scheme, in the sense that in the counterfactual diagram the off-path B is blocked from firing. Off-path blocking is maximal in that it is implemented in the counterfactual scheme.

(iii) Only if X is on and bifurcating, one needs to distinguish the 'direct' path from X to Y (containing 'in-path' events) and the 'indirect' path(s) from X to Y (containing 'off-path' events); by definition, all events that are not in-path in the direct path are off-path. The direct path is, by definition, the shortest path, once all 'redundant' neurons are collapsed on their antecedent 'parent' with the same on/off state; such redundant neurons have, by definition, one connection in and out. For instance, D in Fig. 1 and Fig. 2 is a redundant neuron in this sense; long chains of similar neurons (as in diagram 18 in Tables 1 and 2) can thus be made shorter. (In very rare cases, the notion of 'shortest path' is ambiguous, since two or more paths could qualify as shortest; then one can choose any of those paths as shortest/direct path: see below.)

---

[12] The clause is then automatically satisfied as a consequence of the 'implementation' specified under (i), see also below.

If the X-neuron under scrutiny is on and bifurcating, we *explicitly* need the 'off-path maximal blocking' clause: we need to implement ¬X while *maintaining* X's blocking off-path, i.e. in any segment of all indirect paths. This allows for instance for identifying the firing of C as a cause of the firing of E in Fig. 1, as one would spontaneously do: the off-path factually blocked B is maintained blocked in the counterfactual scenario. (For other examples, see Ex2, Ex3, Ex5 below.) Thus, DEF-1 now stipulates, more precisely, to implement ¬X in the direct path and in all indirect paths, while *maintaining* off-path (so in any indirect path) any blocking, *including* blocking that can be retraced to the (factually firing) X. This is somewhat sophisticated, because it boils down to not fully implementing ¬X in the counterfactual scenario; off-path one needs to maintain any blocking in the indirect path that exists in the factual scenario, also if it stems from the factual firing of X. In sum, in case the X-event under scrutiny corresponds to a bifurcating and firing X-neuron, we now explicitly need the 'off-path under max blocking' clause. The blocking in the indirect path is now maximal in that it is maintained from the factual firing of X. (Recall from (ii), if X is off, off-path blocking is maximal in that it is *implemented* in the counterfactual scheme.)

In very rare cases, we need, in principle, one more clause in specification (iii): If X is on and bifurcating but one cannot define a direct/shortest path according to the definition given, i.e. if there is more than one 'shortest path' starting at X, then one can choose any of those paths as shortest/direct path; the result does not depend on the choice. This is the case for the C-neurons of diagrams 5, 6, 7, 9: they are bifurcating but the two paths towards the last neuron are equal in length. For an explicit application of DEF-1 in this case, see example Ex1 below.

With these specifications, DEF-1 leads in all diagrams of Table 1 (and subsequent tables) discussed in (Paul and Hall 2013) to causes that are considered in this work as the intuitive causes. Clearly, one will note that DEF-1 shares commonalities with several ingredients of existing counterfactual accounts of causation (see e.g. passages in Paul and Hall 2013, p. 20, 21, etc.). But these ingredients have not been combined in the way here proposed. DEF-1 might seem complex, due to the specifications (i)-(iii), but the synthetic expression "¬X (ceteris paribus, off-path under max blocking) implies ¬Y" allows for an easy use once one is accustomed to the implicit notions. This requires some practice, so let us illustrate the definition in a few non-trivial cases, notably involving bifurcating neurons.

**Ex. 1**. **Diagram 5**, an example of double prevention (Paul and Hall 2013, Ch. 5). For this diagram DEF-1 stipulates the following. A+ and F- are causes of E+ by the simple counterfactual rule in (ii) above. C is bifurcating but its two forward paths are equal in length; whichever path one chooses as shortest/direct, F remains off in the counterfactual C- scenario. Hence, according to specification (iii), C+ is *not* a cause of E+ (C- does not lead to E-). Finally, D+ is a cause of E+ by the simple counterfactual rule of (ii) above. These verdicts coincide with those privileged by the experts, even if this diagram is intensely debated (cf. Paul and Hall 2013, p. 198, p. 216, p. 224, p. 247). Since we derive our results from a comparatively broad definition, our model validates the intuition of several philosophers (e.g. Hall 2000, Hitchcock 2001, Paul and Hall 2013, Ch. 5) that transitivity is violated in certain neuron diagrams[13]. Other well-known accounts struggle with this case (Paul and Hall 2013, p 82ff).

**Ex. 2**. **Diagram 8**. According to DEF-1, C+ is a cause of E+, because C- implies E- if the off-path B and G are kept blocked (cf. the blocking stipulation (iii)). This corresponds to the expert verdict (cf. Paul and Hall 2013, p. 88).

**Ex. 3**. **Diagram 10**. According to DEF-1, C+ is a cause of E+, because C- implies E- if the off-path D is kept blocked (in accordance with (iii)). A+ is a cause of E+ because A- leads to E- (the off-path D is kept blocked by C). These results agree with Paul and Hall (2013, p. 90). Again many other models have been proposed, but none appears to have a wide applicability.

**Ex. 4**. **Diagram 12**, an example of double prevention. DEF-1 implies that A+ and C+ are both causes of E+, due to the simple counterfactual rule applicable to non-bifurcating neurons. Paul and Hall converge to a similar conclusion, even if its justification is again highly debated (Paul and Hall 2013, p. 202).

**Ex. 5**. **Diagram 15**. According to DEF-1, C+ is a cause of E+, because C- implies E-: the off-path B must be kept blocked. A+ is a cause of E+ because of the simple counterfactual rule. The final verdict by Paul and Hall is the same (2013, p. 199).

---

[13] Because DEF-1 implies that in diagram 5 C+ is a cause of D+, which is a cause of F-, which is a cause of E+. Yet C+ is not a cause of E+, as again follows from the model.

Ex. 6. **Diagram 17**, a case of 'redundant prevention'. At t1 only C+ is a cause of G-: C- leads to G+ if the off-path B is kept blocked, as it should be ((iii)). Compare to the discussion by Paul and Hall (2013, p. 213).

Ex. 7. **Diagram 18**. C+ is a cause of I+: C- leads to I- if the off-path B is kept blocked, as it should be ((iii)). Compare to the discussion by Paul and Hall (2013, p. 220).

Let us emphasise that DEF-1 does not work for all neuron diagrams discussed in (Paul and Hall 2013). It cannot work for diagrams that need more or different rules than the 'classic' diagrams, i.e. those that are based on the rules given for diagram 1 in Fig. 1 (plus the rule for double-border neurons as in diagram 10). But we believe that it works for all classic diagrams, perhaps with a minimal upgrading, as further research should confirm[14]. Indeed, we found one diagram in (Paul and Hall 2013) that is classic in our sense, and that might require a slight upgrade of DEF-1 to be applicable also to this case, namely the diagram in Fig. 38, p. 187. On the other hand, this case is ambiguous and there is no final verdict given by the authors regarding the cause(s) of E+. If one wants to identify C+ as a cause in this diagram, then one would need to modify specification (iii) of DEF-1 in following way: "If X is on and bifurcating, …" must be replaced by "If X is on and bifurcating, or on and directly stimulating such an on and bifurcating neuron,…". Interestingly, ChatGPT(4) correctly identified C+ as a cause[15].

Sure, we do not (yet) advocate to use LLMs as always reliable causal experts in cases of philosophical relevance, but in the next section we argue there are reasons to be optimistic as to their future capabilities (beyond the obvious parametric upscaling argument).

## 4. Discussion of the causal tests. Lines of further research

As a preliminary note, let us emphasise that neural diagrams do not necessarily capture all aspects of causation (as also acknowledged by Paul and Hall 2013). To what extent exactly the counterfactual analysis studied via neuron diagrams overlaps with other theories of causation, such as the functional model, the manipulability and the regularity model, remains an open problem. Nevertheless, neuron diagrams cover a large part of causation research in philosophy; and for our task of developing a systematic test they seem an appropriate tool.

---

[14] Cf. footnote 11.
[15] After somewhat lengthy and in principle redundant step-by-step reasoning, ChatGPT(4) concluded about the diagram in (Paul and Hall 2023, Fig. 38, p. 187): "Therefore, E does occur at t4, and its occurrence is directly caused by the sequence of events initiated by C at t1, leading to C1 at t2, then D at t3, which directly causes E to occur at t4."

Are the results obtained by ChatGPT(4) (Tables 1 and 2), DeepSeek-R1, Gemini 2.0 Flash, and o3-mini (Table 3) surprising? We gladly admit that we were surprised that ChatGPT(4) did so well, after we had done tests with ChatGPT(3), which essentially failed on almost all questions. The more recent DeepSeek-R1 and Gemini 2.0 Flash seem to corroborate this tendency of increasing fluency (cf. Table 3). Even a global success rate of say 50% of fully correct answers in Table 1 (and up to 90% in Table 3 for Gemini 2.0 Flash, if one includes partially correct answers) seems an achievement for a technology that works, at its basis, by statistical text completion, and that is not specifically developed for causal reasoning. We suspect that this success rate is difficult to match by humans, on average, and perhaps even by philosophers (but we didn't perform large-scale statistical tests with humans, which seem particularly laborious: they need to be parametrised by a variety of conditions). Unfortunately, it is difficult to get any deep insights into how the LLMs come to their causal verdicts; we are dealing with a black-box technology[16]. Some will maintain, independently of any test results, that this type of technology does not really reason causally (which would be in agreement with the conclusions of e.g. Jin et al. 2023, Zečević et al. 2023); for a critical philosophical assessment along these lines of GPT-4's 'reasoning', see (Arkoudas 2023, 2023a). We indeed notice that when causal diagrams become complex (say, diagrams 19 and 22-25, which are not discussed in the literature, for that matter), the probability of failing becomes much higher, for all chatbots tested. On the other hand, our question Q-CAUSE often does elicit textual answers that are indistinguishable from answers resulting from 'real' step-by-step reasoning – as is known to be the case for other reasoning types (e.g. Chen et al. 2023). From a practical point of view, the conclusion that matters most for us is that several of the tested LLMs can already provide correct answers to causal questions that are considered subtle by the expert community (Tables 2 and 3). Therefore we believe it is far from implausible that next generation LLMs, or dedicated AI, will produce correct answers also for more complex diagrams – equaling or surpassing human experts.

In this context, following observation seems to warrant optimism: there seems to be a relevant analogy between how we constructed DEF-1, and how (future) AIs could identify causes, and perhaps ultimately abstract a definition: the keyword is correlation. We tried – in a somewhat tedious process – various definitions by inspecting large ensembles of neuron diagrams (Table 1), and by looking for patterns in these diagrams, i.e. by looking for correlations. That should be a procedure in which dedicated AIs based on artificial neural nets excel. At any rate, neural nets (which, surely anecdotally, have some resemblance with neuron diagrams) can probably be trained for causal identification in neuron diagrams and in other contexts, and then be used for it. In sum, this analogy in heuristic procedure seems to be one more reason why we believe the 'causal thinking' of (future) AI must be taken seriously, and why the foundations of causation and of artificial neural nets should be a fertile field of interdisciplinary research.

---

[16] The LLMs we tested are, at least in the computer science community, widely accredited with 'step-by-step reasoning' capacities. The detailed algorithmic apparatus to achieve such 'reasoning' is proprietary, but the DeepSeek team's report (Guo et al. 2025) is helpful in disclosing the broad approach, and mentions such techniques as transformer-based neural architecture, attention mechanisms, reinforcement learning, supervised fine-tuning, logical reasoning techniques, etc. We found it also helpful to interact with the DeepSeek chatbot about this question. Finally, let us mention that Buckner (2024) exposes the philosophical (empiricist) pedigree of several of these techniques.

However, as said, our small-scale statistics is not presented here as a measure of whether the studied LLMs are good causal reasoners or not; for that we refer to the specialised literature (e.g. Gao et al. 2023, Jin et al. 2023, Kiciman et al. 2023, Tu et al. 2023, Zečević et al. 2023, Liu et al. 2024).

Before concluding, let us discuss how the test method could be enriched and elaborated. Each of the following routes could be systematically explored in further research projects:

(i) Use the diagrams to test for counterfactual reasoning and mastery of concepts as 'intervention'. It is widely accepted that causal identification is related not only to counterfactual reasoning but also to the concept of intervention (e.g. Woodward 2005). One can indeed use the diagrams to further test whether LLMs master these concepts, as illustrated in Appendix 3, Table 4 by a mini-test on Gemini 2.0 Flash. As for the other tests, we find the answers often excellent, e.g. for diagram 17, of intermediate complexity (but not all answers are fully correct). Again, even if we did not do a large-scale test, one is tempted to conclude that some LLMs 'imagine' causal scenarios as humans do; more precisely, that their answers often coincide with those based on counterfactual / imaginative thinking. Note that when the LLM gives a correct answer (20 out of 25 answers related to the 9 diagrams in Table 4), in appearance it 'understands' counterfactual intervention as humans normally do, notably by keeping all events anterior to the intervening cause fixed. This corresponds to correctly implementing the 'ceteris paribus' clause in DEF-1.

(ii) Paraphrase the prompts. One might wonder how robust the LLMs are under prompt variation. The prompts submitted to the LLMs in Tables 1-4 describe the causal dynamics of the diagrams in detail, carefully representing the time details etc. One can paraphrase these queries by using a more contracted description as is done in typical philosophical phrasings of the diagrams. For instance, the prompt corresponding to the paradigmatic diagram 1 (Fig. 1) can be rephrased thus: "Suppose a scenario in which C occurs and causes D to occur, which subsequently causes E to occur. Suppose A normally causes B to occur, unless C causes D to occur. If B would have occurred, it would have caused E to occur. Does E occur in this scenario? What is/are the cause(s) of E's occurring?" We did a small-scale test with Gemini 2.0 Flash (Thinking Experimental) and ChatGPT o3-mini on the 9 diagrams of Tables 2 and 4. In this test all prompts were rephrased according to the logic as just given for diagram 1. The results showed only small deviations from the results obtained with the original prompts used in Table 3. In some detail, Gemini 2.0 gave one fully wrong answer on diagram 15 and a partially wrong answer for diagram 12 (the answers were correct with the original phrasings used for Table 3). ChatGPT o3-mini gave a fully wrong answer for diagram 17 and partially wrong answers for diagrams 12 and 18; its answers were also problematic for these diagrams under the original prompt (cf. Table 3). We further introduced more abstract phrasings (unlikely to be encountered in the texts used for model training), by replacing in the prompts A, B etc. by Aness, Bness etc., but this led to identical results. In conclusion, these preliminary tests (see also (iii)) suggest that the mentioned LLMs are quite stable under prompt variation for this type of causal reasoning.

(iii) Replace the diagrams by concrete situations, i.e. ask questions about concrete scenarios of which the diagrams are abstract models. For instance, the prompt for diagram 12 (a case of double prevention) could be replaced by following concrete text (paraphrasing an example given by Paul and Hall 2013, p. 175; in square brackets are the corresponding neurons in the diagram): "Bob makes coffee [A], and fills his cup [E]. Meanwhile, Alice scoops up Billy the cat [C] as he lashes his tail wildly [B]; her quick action prevents a disastrous spilling [D], so that the cup remains filled [E]". The test question could then be: What are the causes of the presence of coffee in the cup? Or: What are the causes of the coffee being/remaining in the cup? As another example, diagram 1 in Fig. 1 could be replaced by following description, a concrete example of early preemption (due to Hitchcock 2007): "A poisons V's coffee. V drinks it and dies. If A hadn't poisoned the coffee, B would have, and V would have died anyway. V would not have died if there had been no poison in the coffee. What is/are the cause(s) of V's death?" For what it is worth, we judged that ChatGPT(4) gave correct (and interesting) answers for both problems, well in line with the experts' analysis[17]. Of course, in this case the chatbot can rely on much published text.

---

[17] For instance, to the question what the cause(s) of V's death are, ChatGPT(4) answered (we abbreviate): "1. **Actual Cause**: A's action of poisoning the coffee is the direct or "actual" cause of V's death, because it is the action that directly led to the outcome. In legal and philosophical terms, this is often referred to as the "but-for" cause: […]. 2. **Potential Cause**: B's intention to poison the coffee, should A have failed to do so, introduces a hypothetical alternative cause. B's intent could be considered a potential cause but it did not materialize into the actual cause because A's action preempted B's. Philosophically, this can be viewed through the lens of "counterfactual dependence," suggesting that […] 3. **Proximate

(iv) Complicate the prompts, by generating more complex diagrams, e.g. by using computer code. Since DEF-1 can also be programmed, this could, in principle, lead to a systematic (benchmark) test for this type of causal reasoning. Our preliminary tests suggest that the causal reasoning of existing LLMs breaks down under sufficient diagram complexity. Perhaps it is interesting, notably in an interdisciplinary effort with computer scientists, to do a systematic study of the test score of an AI as a function of diagram complexity. For diagram complexity several numerical measures could be defined; the simplest ones are the number of neurons and the number of columns, i.e. time steps, in the diagram; others are the number of forks, of blocked neurons, of crossings, etc. In the most interesting scenario this would be an easy, quantitative way to get insight into the general depth-of-reasoning of given AIs. For assessing an AI's answers in the case of complex diagrams, an encompassing definition is necessary, one that can be programmed, such as DEF-1.

## 5. Conclusion

In the AI community there is a popular belief that making AI 'think causally' might be a key paradigm for its future development, and perhaps even for reaching AGI (e.g. Pearl 2019, Zečević et al. 2023); a belief also held by philosophers (Buckner 2024, p 74). Several types of causal reasoning tests have been developed by computer scientists and others, in order to monitor and quantify the progress. At the same time, defining the concept of cause is an eminently philosophical topic. Even alone for this reason, an investigation of possible cross-disciplinary fertilization seems in place. Here we proposed a method for testing abstract causal reasoning that is directly inspired by philosophical expertise, gathered since about fifty years (compiled in Paul and Hall 2013), and condensed in neuron diagrams. We showed that the method can well be applied to LLMs as ChatGPT, DeepSeek and Gemini 2.0 Flash. We noted that these LLMs can sometimes give correct causal verdicts in scenarios the interpretation of which is debated in the literature (e.g. Table 2). Now, in order to evaluate results reliably and possibly automatically by a computer algorithm, the method requires an objective standard to compare the AI's answers to. Hence we proposed a definition of cause that works for all diagrams of Table 1, and that might be the basis for an encompassing definition, as further research should establish. Thus, this investigation, which started as

---

Cause**: In legal terms […] In summary, in this scenario, A is both the actual and proximate cause of V's death. B remains a potential but unrealized cause. Thus, in both a philosophical and a legal context, A's action would generally be considered the cause of V's death. B's intention, while relevant to questions of moral or criminal liability if discovered, does not alter the causation of V's death as it did not transition into an action that resulted in harm."

a curiosity-driven inquiry into the reasoning capacities of ChatGPT – making headlines in mass media and hailed by some computer scientists as the beginning of AGI (e.g. Bubeck et al. 2023) –, resulted in a new important philosophical result: namely a definition having a wider applicability than those published hitherto.

Based on the research presented here, and in view of the foreseeable development of ever more powerful AIs, we believe philosophers should take the construction of encompassing definition(s) of cause to heart. We need to be able to verify AI-generated outcomes, especially in subtle causal situations – until now considered the preserve of philosophy. Otherwise, the intellectual authority might soon change sides. This issue has not only intellectual importance, but also practical and ethical, essentially for reasons of trustworthiness. As we emphasized, caution is needed when assessing the 'reasoning' of an AI; and larger-scale experiments are needed to come to definite conclusions. Yet, we believe that the present investigation already hints to the idea that future research in philosophy, even theoretical philosophy, might well profit from, and indeed need, interaction with AI.

## Appendix 1. ChatGPT(4) tested with 25 causal reasoning cases

In the following Table 1 we summarise the results of our causal reasoning test on ChatGPT(4) (see explanations in Section 2). The 25 diagrams depicted here are also the basis for other tests shown in Tables 2, 3, 4.

| Diagram | Correct answer | | Diagram | Correct answer |
|---|---|---|---|---|
| **1.** (From PH, Fig. 1, p. ix)<br> | Yes.<br>C+(t1); D+(t2) | | **14.**<br> | Yes.<br>A+(t1); A1+(t2); F-(t3); A2+(t3) |
| **(2.)** | Yes.<br>C+(t1); D+(t2), B-(t2) | | **15.** (From PH, Fig. 42, p. xx) | Yes.<br>A+(t1), C+(t1); etc. |

| | | | |
|---|---|---|---|
|  | |  | |
| **3.**  | No.<br>C-(t1); D-(t2) | **16.**  | Yes.<br>A+(t1), G+(t1);<br>A2+(t3), F+(t3) etc. |
| **4. (From PH, Fig. 2, p. ix)**  | Yes.<br>A+(t1); B+(t2) | **17. (From PH, Fig. 43, p. xxi)**  | No.<br>C+(t1); D+(t2);<br>E+(t3) |
| **5. (From PH, Fig. 10, p. xi)**  | Yes.<br>A+(t1); D+(t2); F-(t3) | **18. (From PH, Fig. 44, p. xxi)**  | Yes.<br>H+(t1), C+(t1);<br>H3+(t4), G-(t4); etc. |
| **6.**  | No.<br>A-(t1) etc. | **19.**  | No.<br>F+(t1), C-(t1), A-(t1);<br>G+(t4); etc. |
| **7.**  | No.<br>A-(t1); D+(t2); F-(t3); A2-(t3) | **20.**  | No.<br>C+(t1); F+(t3); etc. |
| **8. (From PH, Fig. 12, p. xii)** | Yes.<br>C+(t1); D+(t2) | **(21.)** | Yes. |
| A   B<br>C  D  E<br>F  G | | A   E<br>G  B  F<br>C  D | A+(t1), G-(t1), C-(t1);<br>A2+(t3), F+(t3); etc. |
| **9.**<br>A  B<br>C  D  E<br>F  G | No.<br>C+(t1); B-(t2), G-(t2) | **22.**<br>H   I<br>F   G<br>C  D  E<br>A  B | No.<br><br>H2-(t3); H3-(t4) |
| **10.** (From PH, Fig. 14, p. xiii)<br>A   B<br>D   E<br>C  F | Yes.<br>A+(t1), C+(t1);<br>B+(t2), F+(t2) | **23.**<br>H   I<br>F   G<br>C  D  E<br>A  B | Yes.<br>F+(t1), C+(t1);<br>F1+(t2); E+(t3); G-(t4) |
| **11.**<br>A   B<br>D   E<br>C  F | No.<br>A-(t1) | **24.**<br>H   I<br>F   G<br>C  D  E<br>A  B | Yes.<br>F+(t1) and effects |
| **12.** (From PH, Fig. 29, p. xvii)<br>A   E<br>B  D<br>C | Yes.<br>A+(t1); C+(t1) | **25.**<br>H   I<br>F   G<br>C  D  E<br>A  B | **Yes.**<br>F+(t1); F1+(t2);<br>H2+(t3), E+(t3);<br>H3+(t4), G-(t4) |
| **13.**<br>A   E<br>G  B  F<br>C  D | No.<br>G+(t1), C-(t1);<br>B+(t2); F+(t3) | | |

**Table 1**. All diagrams of the causal test, in which we asked ChatGPT(4) whether the last neuron in the diagram fires or not, and what the causes of the firing or not firing are. In column 2 and 4 the correct answers are given according to DEF-1 (when no confusion exists we do not always mention all causes, cf. main text). For examples of ChatGPT(4)'s full answers, see Table 2. Here we only indicate whether ChatGPT(4) gave incomplete or wrong answers by underlining the diagram number, as in diagram **3**, **5**, etc. Partially correct answers are indicated with extra brackets, as in **(2)**. Several diagrams are discussed in (Paul and Hall 2013) (PH in the table).

## Appendix 2. Comparison between advanced versions of ChatGPT, DeepSeek, Gemini

In the following Table 3 we compare DeepSeek-R1, Gemini 2.0 Flash (Thinking Experimental) and ChatGPT o3-mini. The diagrams and the questions are the same as in Table 1, except that (i) we only ask to give the causes *at the earliest time* (t1 normally), and (2) we specify in the prompt that the answers should be short. In detail, the prompt for diagram 1 was:

"Suppose time t1 is earlier than time t2, which is earlier than time t3. If C would occur at t1, D would occur at t2. If D would occur at t2, E would occur at t3. If A would occur at t1, B would occur at t2, unless C would occur at t1. If B would occur at t2, E would occur at t3. Suppose C and A occur at t1. Does E occur at t3? What is/are the cause(s) at t1 of E's occurring or not occurring? Answer without stating your reasoning steps. So simply state in one or two sentences whether E occurs or not, and what the initial cause(s) (so the cause(s) at t1) of this event are."

All prompts for all other diagrams follow the same logic. "Correct answers" (column 2) are given by DEF-1. Best-in-class for this small-scale test was Gemini 2.0 Flash (Thinking Experimental) with 14 fully correct answers, 9 partially correct answers, and 2 wrong answers; followed by DeepSeek-R1 (resp. 10 / 15 / 0). ChatGPT o3-mini had only 2 fully correct and 11 partly correct answers, while 12 answers were wrong, as can be read in the Table.

| # | Diagram | Correct answer | ChatGPT o3-mini | DeepSeek-R1 (DeepThink) | Gemini 2.0 Flash (Thinking Experimental) |
|---|---------|----------------|-----------------|--------------------------|-------------------------------------------|
| 1 |  | Yes. C+(t1) | Yes, E occurs at t3. The causes at t1 of E's occurring <u>are A</u> and C. | E occurs at t3. The initial cause at t1 is the occurrence of C, which leads to D at t2 and subsequently E at t3. | Yes, E occurs at t3. The cause at t1 of E's occurring is C's occurrence at t1. |

| 2 | 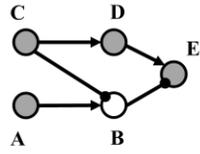 | Yes. C+(t1) | <u>No, E does not occur at t3. The causes at t1 of E's not occurring are A and C.</u> | E occurs at t3. The initial cause at t1 is the occurrence of C, which leads to D at t2, resulting in E at t3. | Yes, E occurs at t3. The cause at t1 of E's occurring is C. |
| 3 | 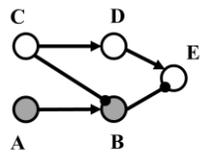 | No. C-(t1) | <u>Yes, E occurs at t3. The causes at t1 of E's occurring are A.</u> | E does not occur at t3. The initial cause at t1 is <u>the occurrence of A</u> (and the absence of C), <u>which leads to B at t2.</u> | No, E does not occur at t3. The causes at t1 of E's not occurring <u>are A</u> and the absence of C. |
| 4 | 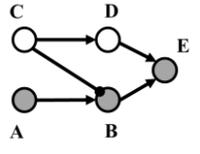 | Yes. A+(t1) | Yes, E occurs at t3. The causes at t1 of E's occurring are A. | E occurs at t3. The initial cause at t1 is the occurrence of A <u>and the absence of C.</u> | <u>No, E does not occur at t3. The causes at t1 of E's not occurring are A and the absence of C.</u> |
| 5 | 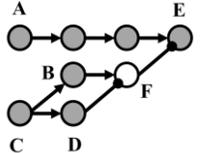 | Yes. A+(t1) | <u>No, E does not occur at t4. The causes at t1 of E's not occurring are A and C.</u> | E occurs at t4. The initial causes at t1 are the occurrence of both A <u>and C.</u> | Yes, E occurs at t4. The cause at t1 of E's occurring is A. |
| 6 | 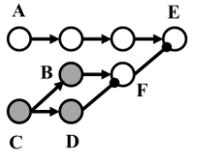 | No. A-(t1) | No, E does not occur at t4. <u>The cause at t1 of E's not occurring is C.</u> | E does not occur at t4. The initial cause at t1 is the <u>occurrence of C</u> and the absence of A. | No, E does not occur at t4. The cause at t1 of E's not occurring is the non-occurrence of A. |
| 7 | 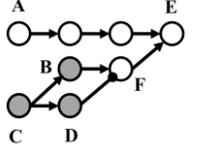 | No. A-(t1) | No, E does not occur at t4. <u>The cause at t1 of E's not occurring is C.</u> | E does not occur at t4. The initial cause at t1 is the <u>occurrence of C</u> and the absence of A. | No, E does not occur at t4. The cause at t1 of E's not occurring is the non-occurrence of A. |
| 8 | 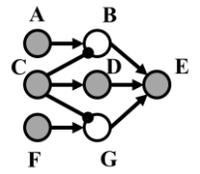 | Yes. C+(t1) | Yes, E occurs at t3. The causes at t1 of E's occurring are <u>A</u>, C, <u>and F.</u> | E occurs at t3. The initial cause at t1 is the occurrence of C. | Yes, E occurs at t3. The cause at t1 of E's occurring is C. |
| 9 | 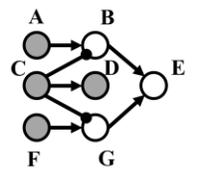 | No. C+(t1) | No, E does not occur at t3. The causes at t1 of E's not occurring are <u>A</u>, C, <u>and F.</u> | E does not occur at t3. The initial cause at t1 is the occurrence of C. | No, E does not occur at t3. The cause at t1 of E's not occurring is C. |

| 10 | 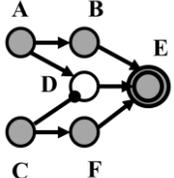 | Yes. A+(t1), C+(t1) | <u>No, E does not occur at t3. The causes at t1 of E's not occurring are A and C.</u> | E occurs at t3. The initial causes at t1 are the occurrence of both A and C. | Yes, E occurs at t3. <u>The cause at t1 of E's occurring is A.</u> |
|----|---|---|---|---|---|
| 11 | 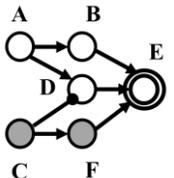 | No. A-(t1) | No, E does not occur at t3. <u>The cause at t1 of E's not occurring is C.</u> | E does not occur at t3. The initial cause at t1 is the <u>occurrence of C</u> and the absence of A. | No, E does not occur at t3. The causes at t1 of E's not occurring are the non-occurrence of A <u>and the occurrence of C.</u> |
| 12 | 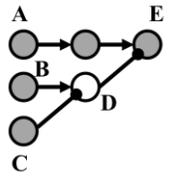 | Yes. A+(t1) C+(t1) | <u>No, E does not occur at t3. The causes at t1 of E's not occurring are A, B, and C.</u> | E occurs at t3. The initial causes at t1 are the occurrence of A and C. | Yes, E occurs at t3. The causes at t1 of E's occurring are A and C. |
| 13 | 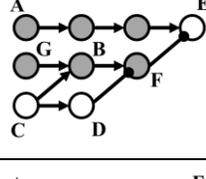 | No. G+(t1), C-(t1) | No, E does not occur at t4. The causes at t1 of E's not occurring <u>are A</u> and G. | E does not occur at t4. The initial causes at t1 are the occurrence of <u>A</u> and G. | No, E does not occur at t4. <u>The cause at t1 of E's not occurring is G.</u> |
| 14 | 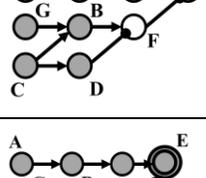 | Yes. A+(t1), C+(t1) | Yes, E occurs at t4. The causes at t1 of E's occurring are A and C. | E occurs at t4. The initial causes at t1 are the occurrence of A and C. | Yes, E occurs at t4. The causes at t1 of E's occurring are A and C. |
| 15 | 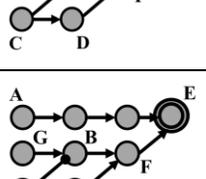 | Yes. A+(t1), C+(t1) | Yes, E occurs at t4. The causes at t1 of E's occurring are A, <u>G,</u> and C. | E occurs at t4. The initial causes at t1 are the occurrence of A and C. | Yes, E occurs at t4. The causes at t1 of E's occurring are A and C. |
| 16 | 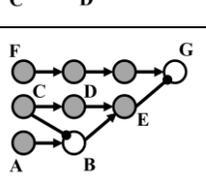 | Yes. A+(t1), G+(t1) | <u>No, E does not occur at t4. The causes at t1 of E's not occurring are A and G.</u> | E occurs at t4. The initial causes at t1 are the occurrence of A and G. | Yes, E occurs at t4. The causes at t1 of E's occurring are A and G. |
| 17 | 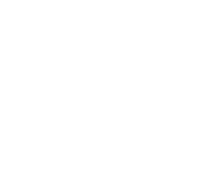 | No. C+(t1) | No, G does not occur at t4. The causes at t1 of G's not occurring are <u>F,</u> C, <u>and A.</u> | E does not occur at t4. The initial causes at t1 are the occurrence of <u>F</u> and C. | No, G does not occur at t4. The cause at t1 of G's not occurring is C. |

| # | | | | | |
|---|---|---|---|---|---|
| 18 |  | Yes. H+(t1), C+(t1) | No, I does not occur at t5. The causes at t1 of I's not occurring are H, F, C, and A. | I occurs at t5. The initial causes at t1 are the occurrence of H and C. | Yes, I occurs at t5. The cause at t1 of I's occurring is H. |
| 19 |  | No. F+(t1), C-(t1), A-(t1) | No, I does not occur at t5. The causes at t1 of I's not occurring are H and F. | I does not occur at t5. The initial causes at t1 are the occurrence of H and F. | No, I does not occur at t5. The cause at t1 of I's not occurring is F. |
| 20 |  | No. C+(t1) | Yes, E occurs at t4. The causes at t1 of E's occurring are A and C. | E does not occur at t4. The initial cause at t1 are the occurrence of A and C. | Final answer: No, E does not occur at t4. The cause at t1 of E's not occurring is C. |
| 21 |  | Yes. A+(t1), G-(t1), C-(t1) | No, E does not occur at t4. The cause at t1 of E's not occurring is A. | E occurs at t4. The initial cause at t1 is the occurrence of A. | Final answer: No, E does not occur at t4. The cause at t1 of E's not occurring is C. |
| 22 |  | No. H2-(t3) | No, I does not occur at t5. The causes at t1 of I's not occurring are H, F, C, and A. | I does not occur at t5. The initial causes at t1 are the occurrence of H, F, C, A. | Final answer: No, I does not occur at t5. The cause at t1 of I's not occurring is H. |
| 23 |  | Yes. F+(t1), C+(t1) | No, I does not occur at t5. The causes at t1 of I's not occurring are H, F, C, and A. | I occurs at t5. The initial causes at t1 are the occurrence of H, F, C, and A. | Final answer: Yes, I occurs at t5. The causes at t1 of I's occurring are H and F. |
| 24 |  | Yes. F+(t1) | No, I does not occur at t5. The causes at t1 of I's not occurring are H, F, C, and A. | I occurs at t5. The initial causes at t1 are the occurrence of H, F, C, and A. | Final Answer: Yes, I occurs at t5. The cause at t1 of I's occurring is F. |

| 25 |  | Yes. F+(t1) | No, I does not occur at t5. The causes at t1 of I's not occurring are H, F, C, and A. | I occurs at t5. The initial causes at t1 are the occurrence of H, F, C, and A. | Final Answer: Yes, I occurs at t5. The causes at t1 of I's occurring are H and F. |

**Table 3**. Three chatbots are asked to (succinctly) reply to the question: Does the last neuron in the diagram fire, and what is/are the initial cause(s)? Wrong parts of the answers are underlined. (If only a part of the initial causes is listed, the answer does not count as fully correct, and the answer is partly underlined.) Correct answers follow from DEF-1. Gemini 2.0 Flash (Thinking Experimental) scores best on this small-scale test, followed by DeepSeek-R1.

## Appendix 3. More tests on counterfactual reasoning

Table 4 shows the results obtained with Gemini 2.0 Flash (Thinking Experimental) on a test for causal and counterfactual reasoning involving the concept of intervention. As explained in the caption of the Table, first the usual question is asked; immediately after the chatbot gives its answer other prompts are submitted. The diagrams are the same as in Table 2, and are all, but one (2), discussed in the literature. The results are discussed in Section 4.

| # | Diagram | Transcription + Questions | Correct answer to the 1st question | Gemini 2.0 Flash (Thinking Experimental) |
|---|---------|---------------------------|-----------------------------------|------------------------------------------|
| 1 |  | Suppose time t1 is earlier than time t2 <…> Does E occur at t3? What is/are the cause(s) at t1 of E's occurring or not occurring? <br><br> **** <br> Suppose now that in the situation just sketched, event D does not occur, because of an external intervention. Does E occur now in this new situation? Why? | Yes. C+(t1) | Yes, E occurs at t3. The cause at t1 of E's occurring is C. <br><br> **** <br> No, E still occurs in this new situation. This is because even though D does not occur, A still occurs at t1, which leads to B at t2, and subsequently B at t2 leads to E at t3. |
| 2 |  | Suppose time t1 is earlier than time t2 <…> Does E occur at t3? What is/are the cause(s) at t1 of E's occurring or not occurring? <br><br> **** <br> Suppose now that in the situation just sketched, event D does not occur, because of an external intervention. Does E occur now in this new situation? Why? | Yes. C+(t1) | Yes, E occurs at t3. The causes at t1 of E's occurring are C and A. <br><br> **** <br> No, E does not occur now in this new situation. This is because the initial condition for E to occur through D (namely D's occurrence) is removed by external intervention, and the alternative condition for E to occur through B (namely B's occurrence) does not happen either because C's occurrence at t1 negates A's influence on B at t2 as per the given rules. |

| 5 | 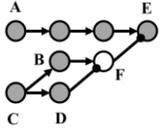 | Suppose time t1 is earlier than time t2 <...> Does E occur at t3? What is/are the cause(s) at t1 of E's occurring or not occurring?<br><br>****<br><br>Suppose now that in the situation just sketched, event D does not occur, because of an external intervention. Does E occur now in this new situation? Why?<br><br>****<br><br>If D does not occur (by an external intervention), E does not occur, because F now blocks E. In this situation, how could you make sure that E occurs by an external intervention? | Yes.<br>A+(t1) | Yes, E occurs at t4. The causes at t1 of E's occurring are A <u>and C.</u><br><br>****<br>No, E does not occur now in this new situation. This is because the non-occurrence of D, due to the external intervention, does not prevent B from occurring (as B's occurrence depends only on C at t1). The occurrence of B at t2 then leads to F occurring at t3, which in turn prevents A2 at t3 from causing E at t4.<br>****<br>To make sure that E occurs, you could intervene at time t2 to prevent event B from occurring. This is because preventing B would stop F from occurring at t3, thus removing the block for E to occur at t4 given A2 occurs at t3. |
| 8 | 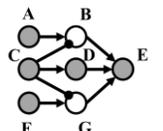 | Suppose time t1 is earlier than time t2 <...> Does E occur at t3? What is/are the cause(s) at t1 of E's occurring or not occurring?<br>****<br>Suppose now that in the situation just sketched, event D does not occur, because of an external intervention. Does E occur now in this new situation? Why?<br><br>****<br><br>If D does not occur by an external intervention, E does not occur. In this situation, how could you make sure that E occurs by an external intervention? | Yes.<br>C+(t1) | Yes, E occurs at t3. The cause at t1 of E's occurring is C.<br><br>****<br>No, E does not occur now in this new situation. This is because the conditions for E to occur via D, B or G at t2 are not met. D does not occur due to external intervention. Because C occurs at t1, neither B nor G occurs at t2. As none of B, D, or G occurs at t2, E does not occur at t3.<br>****<br>To make sure that E occurs, you could intervene at time t2 to make event B occur. This is because if B occurs at t2, then according to the given rule "If B would occur at t2, E would occur at t3", E will occur at t3. |
| 10 | 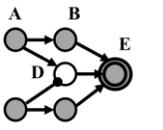 | Suppose time t1 is earlier than time t2 <...> Does E occur at t3? What is/are the cause(s) at t1 of E's occurring or not occurring?<br>****<br>Suppose now that in the situation just sketched, event D does occur, because of an external intervention. Does E occur now in this new situation? Why?<br><br>****<br><br>If D occurs by an external intervention, E still occurs. In this situation (A, D, C occurring), how could you make sure by external interventions that E does not occur? | Yes.<br>A+(t1),<br>C+(t1) | Yes, E occurs at t3. The causes at t1 of E's occurring are A and C.<br><br>****<br>Yes, E still occurs in this new situation. This is because B still occurs at t2 due to A at t1, and with D now occurring at t2 due to external intervention, the condition "if B would occur at t2, E would occur at t3, if also either D at t2 or F at t2 (or both) would occur" is satisfied, leading to E's occurrence at t3.<br>****<br>*<long preliminary reasoning (correct)>*<br>Final Answer: To make sure that E does not occur by external interventions, you should intervene at time t2 to prevent both event B and event F from occurring. |

| 12 | 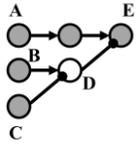 | Suppose time t1 is earlier than time t2 <…> Does E occur at t3? What is/are the cause(s) at t1 of E's occurring or not occurring? | Yes. A+(t1); C+(t1) | Yes, E occurs at t3. The causes at t1 of E's occurring are A and C. |
|---|---|---|---|---|
| | | **** Suppose now that in the situation just sketched, event D does occur, because of an external intervention. Does E occur now in this new situation? Why? | | **** No, E does not occur now in this new situation. This is because although A1 still occurs at t2, event D also occurs at t2 due to the external intervention, and according to the rule "If A1 would occur at t2, E would occur at t3, unless D would occur at t2", the occurrence of D at t2 prevents E from occurring at t3. |
| | | **** If D occurs by an external intervention, E does not occur, because it is blocked by D. In this situation, how could you make sure by an external intervention that E does occur ? | | **** To make sure that E occurs by an external intervention, you should intervene at time t2 to prevent event D from occurring. This counteracts the initial external intervention causing D, and as a result, because A1 still occurs at t2, and D is prevented from occurring, E will occur at t3. |
| 15 | 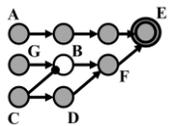 | Suppose time t1 is earlier than time t2 <…> Does E occur at t3? What is/are the cause(s) at t1 of E's occurring or not occurring? | Yes. A+(t1), C+(t1) | Yes, E occurs at t4. The causes at t1 of E's occurring are A and C. |
| | | **** Suppose now that in the situation just sketched, event D does not occur, because of an external intervention. Does E occur now in this new situation? Why? | | **** No, E does not occur now in this new situation. This is because with D not occurring and with C occurring at t1, event F does not occur at t3, which is a necessary condition for A2 at t3 to cause E at t4. |
| | | **** If D does not occur (by an external intervention), E does not occur. In this situation, how could you make sure by external interventions that E does occur ? | | **** To make sure that E occurs by external interventions, you should intervene at time t2 to make event B occur. This is because if B occurs at t2, then F will occur at t3, and since A2 also occurs at t3, the condition for E to occur at t4 will be met. |
| 17 | 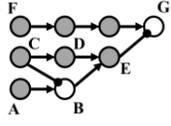 | Suppose time t1 is earlier than time t2 <…> Does G occur at t3? What is/are the cause(s) at t1 of G's occurring or not occurring? | No. C+(t1) | No, G does not occur at t4. The cause at t1 of G's not occurring is C. |
| | | **** Suppose now that in the situation just sketched, event D does not occur, because of an external intervention. Does G occur now in this new situation? Why? | | **** Yes, G occurs now in this new situation. This is because the removal of D by external intervention means that E does not occur at t3. Since E does not occur at an earlier time, F2 at t3 leads to G occurring at t4. |
| | | **** If D does not occur (by an external intervention), G occurs, because E does not occur and cannot block G. In this situation, how could you make sure by an external intervention that G does not occur ? | | **** To make sure that G does not occur by an external intervention, you should intervene at time t2 to prevent event F1 from occurring. This is because preventing F1 at t2 will prevent F2 at t3, and consequently prevent G at t4 as G's occurrence requires F2 at t3 and the absence of E at an earlier time. |

| 18 | 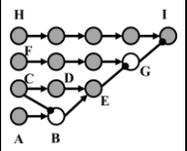 | Suppose time t1 is earlier than time t2 <...> Does I occur at t3? What is/are the cause(s) at t1 of I's occurring or not occurring?<br><br>****<br><br>Suppose now that in the situation just sketched, event D does not occur, because of an external intervention. Does I occur now in this new situation? Why?<br><br>****<br><br>If D does not occur (by an external intervention), I does not occur, because E does not occur, therefore G occurs and prevents I from occurring. In this situation, how could you make sure by an external intervention that I does occur ? | Yes.<br>H+(t1),<br>C+(t1) | Yes, I occurs at t5. The causes at t1 of I's occurring are H and C.<br><br>****<br><br><u>Yes, I still occurs in this new situation. This is because the sequence of events from H at t1 leading to I at t5 operates independently of event D. The occurrence of H at t1 is sufficient for I to occur at t5, regardless of whether D occurs.</u><br><br>****<br><br>*<long preliminary reasoning (largely correct)>*<br>Final answer: To make sure that I does occur by an external intervention, you should intervene at time t2 to make event B occur. |

**Table 4**. Gemini 2.0 Flash (Thinking Experimental) is first asked to (succinctly) reply to the usual question [Does the last neuron in the diagram fire, and what is/are the initial cause(s)?]. Immediately after its answer it is asked a new question, stated after the '****' in column 3; often a third question is asked. (Note that before asking the third question, we succinctly gave the correct answer to the second question.) Wrong parts of the answers are underlined as before. Correct answers to the first question follow from DEF-1.